\documentclass[letterpaper, 10 pt, conference]{ieeeconf}

\IEEEoverridecommandlockouts
\usepackage{cite}
\usepackage{amsmath,amssymb,amsfonts}
\usepackage{algorithmic}
\usepackage{graphicx}
\usepackage{textcomp}
\usepackage[table]{xcolor}
\usepackage{comment}
\usepackage[ruled,vlined]{algorithm2e}
\usepackage[font=small]{caption}
\usepackage{array}
\usepackage{subcaption}
\usepackage{arydshln}

\usepackage{multirow}

\usepackage{tensor}
\newcommand{\eg}{\emph{e.g.},}
\newcommand{\ie}{\emph{i.e.},}

\usepackage[bookmarks=false]{hyperref}
\usepackage{float}
\usepackage{rotating}
\usepackage{nicematrix}  %
\usepackage{mathtools}
\usepackage{amsmath}
\usepackage{pifont}
\usepackage{cuted}
\usepackage{dblfloatfix}
\usepackage{balance}
\usepackage{titlesec}

\usepackage{dsfont}
\usepackage{mathtools}
\usepackage{booktabs}
\usepackage{graphics}
\usepackage[mathscr]{euscript}

\renewcommand{\vec}[1]{{\mathbf #1}}

\newcommand{\cmark}{\ding{51}}%
\newcommand{\xmark}{\ding{55}}%

\newcommand{\algstep}[1]{\item[]\medskip\hrule\kern 2pt\hbox to \textwidth{\hspace{\labelsep}\textbf{#1}\hfill}\hrule}

\usepackage{xspace}
\makeatletter
\DeclareRobustCommand\bmvaOneDot{\futurelet\@let@token\bmv@onedotaux}
\def\bmv@onedotaux{\ifx\@let@token.\else.\null\fi\xspace}
\makeatother

\definecolor{LightCyan}{rgb}{0.88,1,0.88}
\definecolor{linear_color}{RGB}{220,223,240}
\definecolor{gray_bbox_color}{RGB}{243,243,244}
\newcommand{\coolname}{\textit{$SPREAD$}}
\definecolor{rebuttal}{rgb}{0,0,1}

\def\figref#1{Fig.~\ref{#1}}

\def\eqref#1{Eq.~(\ref{#1})}

\newcommand\blfootnote[1]{%
  \begingroup
  \renewcommand\thefootnote{}\footnote{#1}%
  \addtocounter{footnote}{-1}%
  \endgroup
}

\begin{document}

\bstctlcite{IEEEexample:BSTcontrol}

\title{\LARGE \bf SPREAD: Subspace Representation Distillation for \\Lifelong Imitation Learning}

\author{Kaushik Roy$^{1}$, Giovanni D’urso$^{1}$, Nicholas Lawrance$^{1}$, Brendan Tidd$^{1}$, Peyman Moghadam$^{1,2}$
}

\twocolumn[{%
\renewcommand\twocolumn[1][]{#1}%
\maketitle
\begin{center}
    \centering
    \captionsetup{type=figure}
    \vspace{-6mm}
    \includegraphics[width=.9\textwidth]{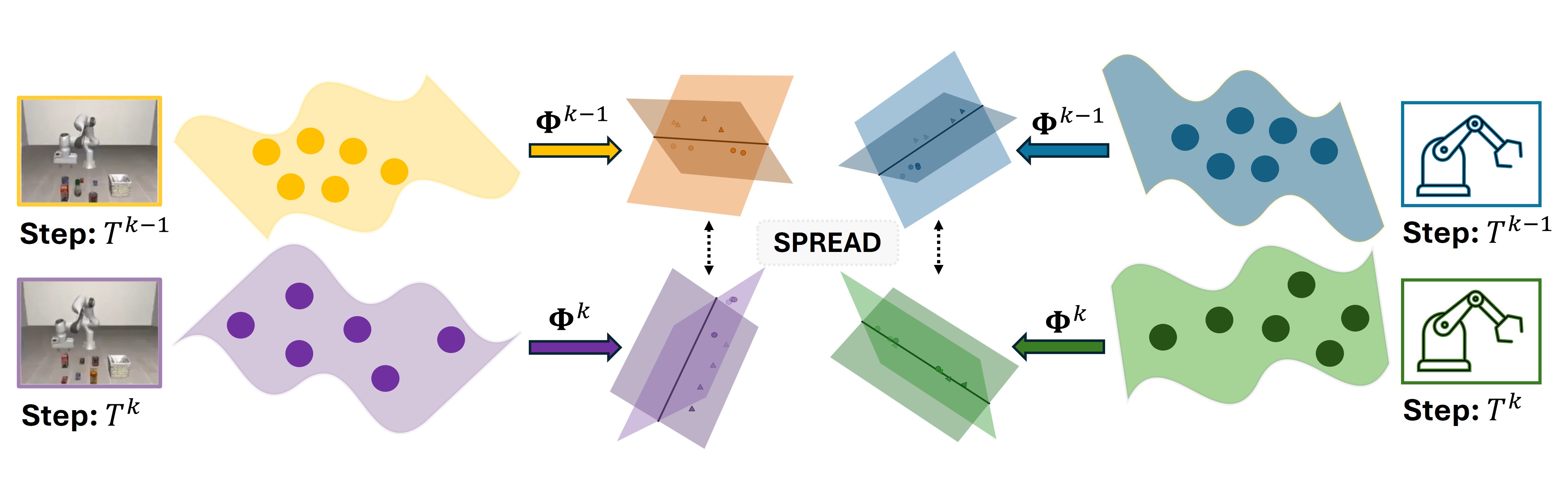}
    \vspace{-1.0em}
    \captionof{figure}{Geometrical interpretation of the Subspace Representation Distillation loss. \coolname{} maximizes similarity between consecutive LIL policies by minimizing discrepancies in projected feature representations across different input modalities within each policy’s subspace. }
    \label{fig:spread}
\end{center}%
}]

\everypar{\looseness=-1}

\begin{abstract}

A key challenge in lifelong imitation learning (LIL) is enabling agents to acquire new skills from expert demonstrations while retaining prior knowledge. This requires preserving the low-dimensional manifolds and geometric structures that underlie task representations across sequential learning. Existing distillation methods, which rely on $L_2$-norm feature matching in raw feature space, are sensitive to noise and high-dimensional variability, often failing to preserve intrinsic task manifolds. To address this, we introduce \textit{\coolname{}}, a geometry-preserving framework that employs singular value decomposition (SVD) to align policy representations across tasks within low-rank subspaces. This alignment maintains the underlying geometry of multimodal features, facilitating stable transfer, robustness, and generalization. Additionally, we propose a confidence-guided distillation strategy that applies a Kullback–Leibler divergence loss restricted to the top-$M$ most confident action samples, emphasizing reliable modes and improving optimization stability. Experiments on the LIBERO, lifelong imitation learning benchmark, show that \textbf{\coolname{}} substantially improves knowledge transfer, mitigates catastrophic forgetting, and achieves state-of-the-art performance.
\blfootnote{$^1$ CSIRO Robotics, DATA61, CSIRO, Australia. 
E-mails: {\tt\footnotesize \emph{firstname.lastname}@csiro.au}}
\blfootnote{$^2$ Queensland University of Technology (QUT), Brisbane, Australia. 
E-mails: {\tt\footnotesize \emph{firstname.lastname}@qut.edu.au}}

\end{abstract}

\section{Introduction}

Robotic agents are increasingly required to operate in open-world environments where new tasks arise sequentially and previously acquired skills must be retained over long horizons. While imitation learning~\cite{hussein2017imitation, schaal1996learning, billard2016learning, stepputtis2020language, xie2024decomposing} has demonstrated strong performance in teaching robots complex behaviors from demonstrations, its standard formulations assume a fixed task distribution and thus fail under continual task acquisition. When exposed to sequential learning, policy networks suffer from catastrophic forgetting~\cite{kirkpatrick2017overcoming, french1999catastrophic}, where adapting to new demonstrations degrades representations required for prior skills. Addressing this challenge is central to enabling scalable and reliable lifelong imitation learning.

A variety of strategies have been proposed to alleviate catastrophic forgetting in lifelong imitation learning. Experience Replay (ER) \cite{chaudhry2019tiny} addresses forgetting by replaying samples from previous tasks, but imbalanced data distributions often shift the representation space toward the current task, degrading performance on earlier tasks. Hierarchical skill learning methods such as LOTUS \cite{wan2024lotus} enhance scalability through a dynamic skill library, but remain vulnerable to degradation of past knowledge. Distillation-based approaches, such as M2Distill \cite{roy2024m2distill}, transfer multimodal knowledge between policies; however, their reliance on $L_2$ feature alignment in raw feature space neglects the underlying geometric structure of task representations.  This limitation can lead to overly rigid alignment that impedes adaptation. These shortcomings highlight the need for subspace-preserving, geometry-aware distillation strategies to enable robust and scalable lifelong imitation learning.

\begin{figure*}[!t]
\vspace{-.5em}
\centering
\includegraphics[width=\textwidth, scale=1]{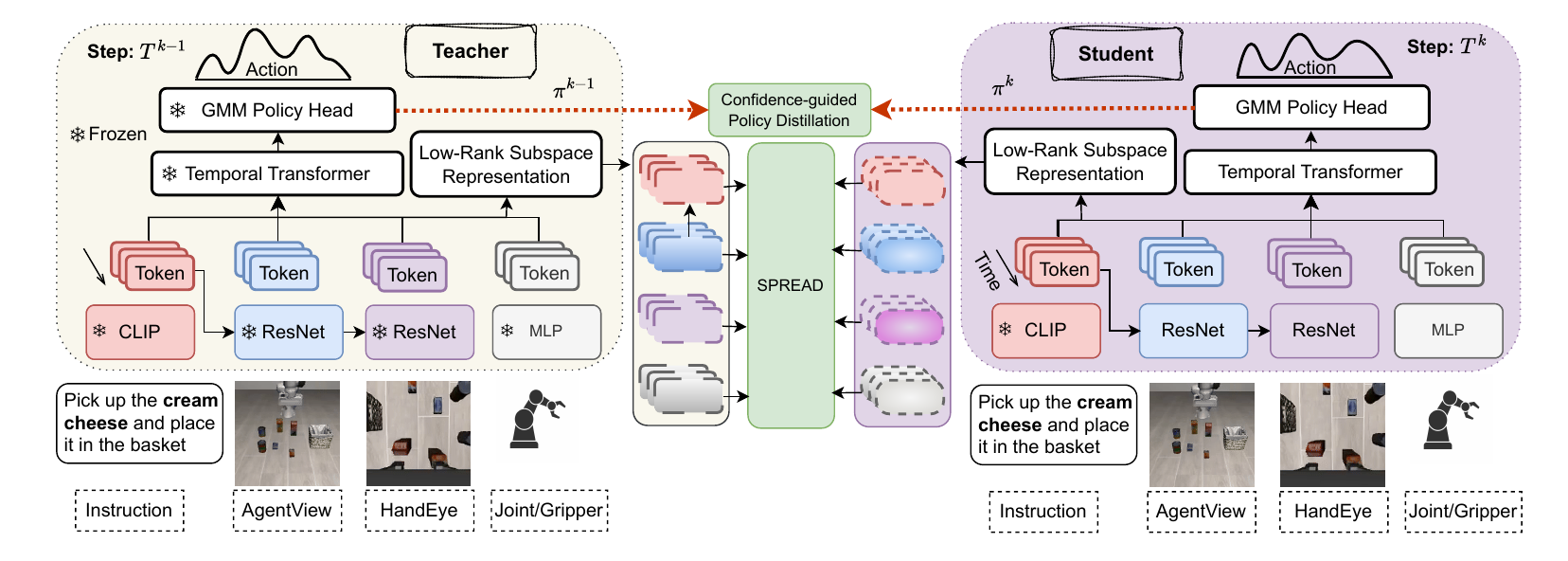}
\vspace{-1em}
\caption{%
Overview of our proposed \coolname{} method. %
Subspace Representation Distillation aligns the latent representations from different input modality encoders (\eg{} Task, AgentView, HandEye, Joint, and Gripper information), while confidence-guided policy distillation maps the action distribution of the GMM policy between incremental steps $T^{k-1}$ and $T^{k}$.
} 
\label{fig:spread++} 
\vspace{-1.0em}
\end{figure*}

To this end, we propose \textbf{\coolname{}}, a lifelong imitation learning framework that explicitly aligns the low-rank subspace representations of consecutive policies. By enforcing consistency in how features are projected onto dominant subspaces, our approach preserves the intrinsic directions that define task manifolds while leaving orthogonal directions available for novel skill acquisition (see Fig.~\ref{fig:spread}). The motivation for subspace alignment arises from the observation that neural representations are often concentrated in low-dimensional subspaces within high-dimensional feature spaces. Prior work in representation learning has demonstrated that these subspaces capture the most discriminative and transferable structure of task-relevant features~\cite{roy2023subspace}, while also exhibiting robustness to noise~\cite{cheng2021nbnet}. Leveraging this property allows \coolname{} to distill knowledge more effectively, avoiding over-constraining raw activations while maintaining the geometry critical for skill retention. This geometry-aware distillation provides a principled balance between stability and plasticity: the representational essence of past tasks is retained, yet sufficient capacity remains to encode new skills without interference. Complementing this, our confidence-guided policy distillation applies Kullback-Leibler (KL) divergence~\cite{kullback1951information} to top-$K$ action samples, ensuring robust behavioral transfer. 

We evaluate \textbf{\coolname{}} on diverse robotic manipulation benchmarks under sequential learning settings. Results show that \coolname{} significantly outperforms conventional feature-level distillation and state-of-the-art lifelong imitation learning baselines. Beyond quantitative gains, our analysis reveals that \coolname{} maintains the intrinsic manifold structure of task features, thereby yielding more robust transfer and retention across tasks.

In summary, this work makes the following contributions:
\begin{itemize}
    \item We introduce \textbf{\coolname{}}, a novel framework that preserves the low-dimensional subspace geometry of task features in lifelong imitation learning. We provide theoretical justification for why subspace-level alignment better preserves intrinsic task manifolds compared to feature-level distillation.
    \item We present a confidence-guided policy distillation strategy, using KL divergence on top-$K$ actions to enhance behavioral robustness.
    \item We demonstrate through extensive experiments that \coolname{} achieves state-of-the-art performance in mitigating catastrophic forgetting, while enabling efficient adaptation to new robotic skills.
\end{itemize}

\label{sec:rel_work}
\section{Related Work}

Imitation learning (IL) enables robots to replicate expert behaviors by mapping observations to actions~\cite{hussein2017imitation,schaal1996learning,billard2016learning, jang2022bc}. Extending IL to sequential settings, Lifelong Imitation Learning (LIL) aims to continuously acquire new skills while retaining past ones, thereby addressing catastrophic forgetting~\cite{french1999catastrophic, kirkpatrick2017overcoming}. Existing LIL approaches draw inspiration from lifelong learning in vision~\cite{shmelkov2017incremental, MAHMOODI2025} and language~\cite{huang2023knowledge}, adapting regularization~\cite{ahn2019uncertainty, roy2023l3dmc}, memory-replay~\cite{rebuffi2017icarl, rolnick2019experience, kirkpatrick2017overcoming, gao2021cril}, and dynamic architecture~\cite{yoon2017lifelong, douillard2022dytox} based strategies to mitigate catastrophic forgetting. Experience Replay (ER)~\cite{chaudhry2019tiny} retains a limited subset of trajectories from past tasks and replays them alongside new demonstrations, but imbalanced replay distributions can bias the learned policy toward more recent skills. Similarly, CRIL~\cite{gao2021cril} leverages deep generative replay (DGR)~\cite{shin2017continual} to generate pseudo-demonstrations through action-conditioned video prediction, however, the instability and inaccuracy of generated videos degrade performance on complex LIL tasks. %
EWC~\cite{kirkpatrick2017overcoming} uses a Fisher information-based measure to constrain updates on parameters essential for earlier tasks to balance plasticity and stability, but struggles to maintain consistent performance in LIL. By decomposing complex behaviors into modular skills, BUDS~\cite{zhu2022bottom} enables effective skill reuse through autonomous segmentation of demonstration trajectories. LOTUS~\cite{wan2024lotus} advances this paradigm by coupling unsupervised skill discovery with open-vocabulary vision models to build a continually expanding skill library. Despite their scalability benefits, both BUDS and LOTUS face challenges in maintaining efficiency and representational consistency as the skill library scales. Tokenization-based frameworks such as T2S~\cite{zhang2025t2s} introduce parameter-efficient skill tokens that extend transformer architectures without uncontrolled parameter growth, while SPECI~\cite{xu2025speci} adopts a prompt-driven skill codebook for efficient skill reuse and adaptation. These approaches improve scalability but often depend on strong architectural assumptions.
Knowledge distillation strategies like PolyTask~\cite{haldar2023polytask} consolidates task-specific policies into a single generalist model. M2Distill~\cite{roy2024m2distill} employs multimodal distillation to align visual, textual, and motor features between old and new policies, thereby reducing modality drift. However, most distillation-based approaches rely on $L_2$ matching in raw feature space, neglecting the intrinsic low-dimensional manifold structure of task features. Our work addresses this gap by introducing \coolname{}, a subspace representation distillation framework that explicitly respects the geometry of task representations.

\section{Proposed Methodology}

\subsection{Problem Formulation}
In lifelong imitation learning (LIL), a robot sequentially learns a policy \(\pi\) over a sequence of tasks \(\{T^1, \dots, T^K\}\) to mimic expert demonstrations while retaining and transferring knowledge across tasks. Each task \(T^k = (\mu_0^k, g^k, l^k)\) is defined by an initial state distribution \(\mu_0^k\), a goal predicate \(g^k: \mathcal{S} \rightarrow \{0, 1\}\), and a language description \(l^k\), within a shared Markov Decision Process \(\mathcal{M} = (\mathcal{S}, \mathcal{A}, \mathcal{T}, H)\), where \(\mathcal{S}\) is the state space, \(\mathcal{A}\) is the action space, \(\mathcal{T}: \mathcal{S} \times \mathcal{A} \rightarrow \mathcal{S}\) is the transition function, and \(H\) is the episode horizon. For each task \(T^k\), the robot receives  \(N\) expert demonstration trajectories \(D^k = \{\tau_i^k\}_{i=1}^N\), along with an associated language description \( l^k \). Each demonstration trajectory \( \tau^k_i \) consists of a sequence of state–action pairs \( \{(s_t, a_t)\}_{t=1}^{L^k} \), where \( L^k < H \) and \( H \) denotes the episode horizon. These demonstrations are typically collected via teleoperation and serve as supervision for training a policy \( \pi \).
The policy is optimized through behavioral cloning~\cite{bain1995framework}, which minimizes the negative log-likelihood of expert actions given the corresponding states:  
\begin{equation}
    \min_{\pi} J(\pi) = \frac{1}{K} \sum_{k=1}^{K} \mathbb{E}_{s_t, a_t \sim D^k} \left[ \sum_{t=1}^{L^k} -\log \pi(a_t \mid s_{\leq t}; T^k) \right].
\end{equation}
A central challenge in this setting is that the robot, while progressing through tasks sequentially, no longer has direct access to demonstrations from earlier tasks. Consequently, the policy must not only sequentially adapt to new tasks but also preserve knowledge from prior tasks, thereby mitigating catastrophic forgetting and enabling effective transfer of skills across the task sequence.

In lifelong imitation learning (LIL), preserving the intrinsic low-dimensional manifold and geometric structure of task features across sequential tasks is critical to mitigate catastrophic forgetting while enabling adaptation to new tasks. Unlike conventional L2-norm-based distillation, which is sensitive to noise and high-dimensional variations, we propose a \textit{subspace representation distillation} loss that aligns low-rank subspace projections of teacher and student features. This approach leverages singular value decomposition (SVD)~\cite{wall2003singular} to capture the principal geometric structure of multimodal features, ensuring robust knowledge transfer. Additionally, for policy distillation, we introduce a confidence-guided Kullback-Leibler (KL) divergence loss that prioritizes the top \( K \) samples with high log probability scores, enhancing stability by focusing on reliable action distributions. \figref{fig:spread++} shows the complete architecture of the proposed \coolname{} framework.

\subsection{Subspace Representation Distillation}

Knowledge distillation~\cite{hinton2015distilling} aims to transfer learned representations from a teacher model to a student model, preserving performance while supporting efficient adaptation. We introduce \textit{\textbf{S}ubs\textbf{p}ace \textbf{Re}present\textbf{a}tion \textbf{D}istillation} (\coolname{}), a geometry-aware distillation method that aligns the principal subspaces of feature representations between teacher and student. Given a feature matrix $ f \in \mathbb{R}^{m \times n} $, we compute its reduced SVD~\cite{eckart1936approximation}: $ f = U \Sigma V^\top $, where $ U \in \mathbb{R}^{m \times r} $ contains the top-$r$ left singular vectors spanning the dominant feature subspace. The projection of $ f $ onto this $r$-dimensional subspace is given by $ U U^\top f $, which captures the most informative directions in the representation.

For teacher and student feature matrices $ f_t $ and $ f_s $, \coolname{} minimizes the discrepancy between their subspace-projected features using the Frobenius norm:
\begin{equation}
\mathcal{L}_{\text{SPREAD}} = \| U_t U_t^\top f_t - U_s U_s^\top f_s \|_F^2 + \| U_t U_t^\top f_s - U_s U_s^\top f_t \|_F^2,
\label{eq:srd_loss}
\end{equation}
where $ U_t $ and $ U_s $ are the top-$r$ left singular vectors of $ f_t $ and $ f_s $, respectively. This symmetric formulation encourages both alignment of the subspace bases (\ie{} $ U_s \approx U_t $) and consistency of feature content within those subspaces. Although subspace representation distillation introduces an $\mathcal{O}(mnr)$ computational overhead, the overhead is limited in practice, and the resulting performance improvements justify the computation.

\smallskip
\noindent
\textbf{Theoretical Significance. }
SPREAD leverages the geometric structure of feature representations by focusing on the principal subspaces, which encapsulate the dominant patterns in the data. The projection \( U U^T f \) retains the most informative components of \( f \), as determined by the largest singular values in \( \Sigma \). By minimizing \( \mathcal{L}_{\text{SPREAD}} \), the student model (at step $k$) aligns its feature subspace with that of the teacher (at step $k-1$), ensuring the preservation of critical task-relevant structures. This approach is robust to noise and model-specific artifacts, unlike direct feature matching (\eg{} \( \| f_t - f_s \|_F^2 \)), as it prioritizes the principal directions of variation. The SPREAD loss further optimizes both the alignment of subspace bases (\( U_s \approx U_t \)) and the content within those subspaces (\( U_s U_s^T f_s \approx U_t U_t^T f_t \)). This can be interpreted as minimizing the principal angles between the column spaces of \( f_t \) and \( f_s \), providing a geometrically sound framework for knowledge transfer. Moreover, SPREAD is invariant to differences in feature magnitude or dimensionality, making it particularly effective for distillation across incremental steps in lifelong imitation learning.

\smallskip
\noindent
In incremental learning, at step $k$, the frozen model from step $k-1$ acts as the teacher, and the model at step $k$ learning task $T^k$ is the student.
For a batch of size $N$ and sequence length $L$, each modality’s latent tensor is reshaped into a $D \times (NL)$ matrix by flattening the batch and temporal dimensions, yielding compact feature matrices for distillation.

\smallskip
\noindent\textbf{Visual Modalities (wrist-mounted (HandEye) and overhead (AgentView) camera).}  
Let $F^{k-1,\epsilon}, F^{k,\epsilon} \in \mathbb{R}^{D \times (NL)}$ denote the reshaped feature matrices extracted using Resnet18~\cite{he2016deep} for modality $\epsilon \in \{\mathrm{HandEye}, \mathrm{AgentView}\}$ at steps $k-1$ and $k$, respectively. The image modality distillation loss is:
\begin{equation}
\mathcal{L}_{\mathrm{image}} = \sum_{\epsilon \in \{\mathrm{HandEye}, \mathrm{AgentView}\}} \mathcal{L}_{\mathrm{SPREAD}}\!\left(F^{k-1,\epsilon},\, F^{k,\epsilon}\right).
\label{eq:loss_image}
\end{equation}

\noindent\textbf{Language (Text).}  
For text features $G^{k-1}, G^{k} \in \mathbb{R}^{D \times (NL)}$ extracted using CLIP~\cite{radford2021learning} language encoder and subsequently passed through an MLP at steps $k-1$ and $k$, we apply subspace representation distillation analogously:
\begin{equation}
\mathcal{L}_{\mathrm{text}} = \mathcal{L}_{\mathrm{SPREAD}}\!\left(G^{k-1},\, G^{k}\right).
\label{eq:loss_text}
\end{equation}

\noindent\textbf{Extra Modalities (\eg{} Joints, Gripper).}  
For proprioceptive signals such as joint angles and gripper state, let $H^{k-1,\epsilon}, H^{k,\epsilon} \in \mathbb{R}^{D \times (NL)}$ denote the reshaped features extracted via an MLP for $\epsilon \in \{\mathrm{joint}, \mathrm{gripper}\}$ at steps $k-1$ and $k$. The distillation loss for these modalities is:
\begin{equation}
\mathcal{L}_{\mathrm{extra}} = \sum_{\epsilon \in \{\mathrm{joint}, \mathrm{gripper}\}} \mathcal{L}_{\mathrm{SPREAD}}\!\left(H^{k-1,\epsilon},\, H^{k,\epsilon}\right).
\label{eq:loss_extra}
\end{equation}

\subsection{Confidence-guided Policy Distillation}
To ensure consistent retention of prior knowledge in lifelong imitation learning, we align the action distributions of previously learned tasks, parameterized as a Gaussian mixture policy. While the Kullback–Leibler (KL) divergence is the standard measure for distributional alignment, a closed-form solution is intractable for mixture models~\cite{hershey2007approximating}; hence, we employ a Monte Carlo approximation. However, uniform sampling introduces variance, as low-probability regions of the prior policy can disproportionately affect estimation and destabilize optimization. To address this, we use high-confidence samples from the prior policy, focusing knowledge transfer on statistically reliable and behaviorally meaningful regions. This selective distillation reduces spurious influence, preserves prior policy structure, and improves the stability of lifelong adaptation.

Let $\pi^{k}$ and $\pi^{k-1}$ be the policies at steps $k$ and $k-1$. For policy distillation, we align action distributions by minimizing a confidence-weighted KL divergence between the current GMM policy \(\pi^k\) and the previous policy \(\pi^{k-1}\). We uniformly sample \(\{\vec{a}^s\}_{s=1}^B \sim \pi^{k-1}\) and select the top \( M = \lfloor 0.9B \rfloor \) samples with the highest log probabilities \(\log \pi^{k-1}(\vec{a}^s)\), indexed by \(\mathcal{S}_M\). The distillation loss is:\begin{equation}
\label{eq:weighted_kl}
\mathcal{L}_{\text{policy}} = \frac{1}{M} \sum_{s \in \mathcal{S}_M} \left( \log \pi^k(\vec{a}^s) - \log \pi^{k-1}(\vec{a}^s) \right),
\end{equation}
Eq.~\ref{eq:weighted_kl} focuses distillation on confident regions of $\pi^{k-1}$ (high density under the prior policy), reducing variance and avoiding misleading gradients from low-likelihood samples.

\smallskip
\noindent
\textbf{Final Optimization Objective.}  
Integrating the task loss with the modality-specific distillation objectives, the final optimization objective updates the policy \(\pi\) by minimizing:

\begin{align}
J(\pi) &= \frac{1}{K} \sum_{k=1}^K \mathbb{E}_{\substack{s_t, a_t \sim D^k \cup \hat{D}^k}} \Bigg[ \sum_{t=0}^{L^k} -\log \pi(a_t \mid (s_t)_{\leq t}; T^k) \notag \\
&\quad + \lambda_i \mathcal{L}_{\text{image}} + \lambda_t \mathcal{L}_{\text{text}} + \lambda_e \mathcal{L}_{\text{extra}} + \lambda_p \mathcal{L}_{\text{policy}} \Bigg].
\end{align}
where $\hat{D}^k$ refers to memory exemplars uniformly sampled from previously learned tasks. $\lambda_i, \lambda_t, \lambda_e,$ and $\lambda_p$ are weighting coefficients that balance the trade-off between stability (preserving past knowledge) and plasticity (adapting to new tasks) during the learning process. This joint optimization takes inspiration from experience replay, to ensure that the policy not only adapts to the current task but also maintains consistency with previously acquired knowledge.

\section{Experimental Settings}
\label{sec:experiments}

\begin{table*}[th!]
\centering
\caption{Experimental results across three different LIBERO task suites. The reported values are averages from three seeds, including the mean and standard error. The best values are highlighted in bold, and the second-best values are underlined. The dash (-) indicates a failure to reproduce results. All metrics are measured based on success rates (\%).}
\label{tab:benchmark}
\resizebox{\textwidth}{!}{%
\begin{tabular}{lccccccccc}
\hline
\multirow{2}{*}{Method} & \multicolumn{3}{c|}{LIBERO-OBJECT}   & \multicolumn{3}{c|}{LIBERO-GOAL}    & \multicolumn{3}{c}{LIBERO-SPATIAL}\\ \cdashline{2-10}
        & FWT ($\uparrow$) & NBT ($\downarrow$) & \multicolumn{1}{c|}{AUC}~($\uparrow$) & FWT ($\uparrow$)       & NBT ($\downarrow$)       & \multicolumn{1}{c|}{AUC}~($\uparrow$) & FWT ($\uparrow$)       & NBT ($\downarrow$)       & AUC~($\uparrow$)      \\ \hline
Sequential               &   62.0 ($\pm$ 1.0)  &  63.0 ($\pm$ 2.0)   &  30.0 ($\pm$ 1.0)                       &      55.0 ($\pm$ 1.0)     &      70.0 ($\pm$ 1.0)     &     23.0 ($\pm$ 1.0)     &  \underline{72.0} ($\pm$ 1.0)   &  81.0 ($\pm$ 1.0)  &  20.0 ($\pm$ 1.0) \\
EWC~\cite{kirkpatrick2017overcoming}                       &   56.0 ($\pm$ 3.0)  &   69.0 ($\pm$ 2.0)  &            16.0 ($\pm$ 2.0)             &      32.0 ($\pm$ 2.0)     &      48.0 ($\pm$ 3.0)     &      6.0 ($\pm$ 1.0)        &  23.0 ($\pm$ 1.0)   &  33.0 ($\pm$ 1.0)  &  6.0 ($\pm$ 1.0) \\
ER~\cite{chaudhry2019tiny}                       &   56.0 ($\pm$ 1.0)  &   24.0 ($\pm$ 1.0)  &            49.0 ($\pm$ 1.0)             &      53.0 ($\pm$ 1.0)     &      36.0 ($\pm$ 1.0)     &      47.0 ($\pm$ 2.0)        &  65.0 ($\pm$ 3.0)   &  27.0 ($\pm$ 3.0)  &  56.0 ($\pm$ 1.0) \\

BUDS~\cite{zhu2022bottom}                        &   52.0 ($\pm$ 2.0)  &   21.0 ($\pm$ 1.0)  &            47.0 ($\pm$ 1.0)             &     50.0 ($\pm$ 1.0)      &      39.0 ($\pm$ 1.0)     &      42.0 ($\pm$ 1.0)        &   -  &  -  &  - \\
LOTUS~\cite{wan2024lotus}                        &   74.0 ($\pm$ 3.0)  &   11.0 ($\pm$ 1.0)  &            65.0 ($\pm$ 3.0)             &    61.0 ($\pm$ 3.0)       &      30.0 ($\pm$ 1.0)     &     56.0 ($\pm$ 1.0)        &   -  &  -  &  -  \\
M2Distill~\cite{roy2024m2distill}                   &   \underline{75.0} ($\pm$ 3.0)  &   \underline{8.0} ($\pm$ 5.0)  &             \underline{69.0} ($\pm$ 4.0)             &       \underline{71.0} ($\pm$ 1.0)    &      \underline{20.0} ($\pm$ 3.0)     &     \underline{57.0} ($\pm$ 2.0)       & \textbf{74.0} ($\pm$ 1.0)   &  \underline{11.0} ($\pm$ 1.0)  &  \underline{61.0} ($\pm$ 2.0)   \\
\rowcolor{green!15} \coolname{} (Ours)                    &   \textbf{81.0} ($\pm$ 2.0)  &   \textbf{8.0} ($\pm$ 1.0)  &             \textbf{73.0} ($\pm$ 3.0)             &       \textbf{78.0} ($\pm$ 1.0)    &      \textbf{9.0} ($\pm$ 5.0)     &     \textbf{72.0} ($\pm$ 1.0)       &  71.0 ($\pm$ 1.0)   &  \textbf{8.0} ($\pm$ 5.0)  &  \textbf{66.0} ($\pm$ 2.0)   \\ \hline
\end{tabular}%
}
\vspace{-1.5em}
\end{table*}

\subsection{Training and Implementation Details}
We train our subspace representation distillation framework on an NVIDIA H100 GPU, employing the same data augmentation strategy and ResNet–Transformer backbone as the ResNet-T baseline~\cite{liu2024libero}. Each incremental task is trained for 50 epochs. To account for varying levels of task complexity, we adopt different regularization weights across the LIBERO suites. For \textbf{LIBERO-OBJECT}, we set $\lambda_{p}=0.003$ and $\lambda_{t}=\lambda_{i}=\lambda_{e}=0.005$. For \textbf{LIBERO-SPATIAL}, the weights are increased to $\lambda_{p}=0.005$ and $\lambda_{t}=\lambda_{i}=\lambda_{e}=0.03$. Finally, for \textbf{LIBERO-GOAL}, due to its higher difficulty, we employ $\lambda_{p}=0.01$ and $\lambda_{t}=\lambda_{i}=\lambda_{e}=0.3$. We fix the subspace rank to $r=48$ for all experiments. We evaluate our method against the following baselines: (1) \textbf{SEQUENTIAL}~\cite{liu2024libero}, which naively fine-tunes tasks sequentially using the ResNet-Transformer; (2) \textbf{EWC}~\cite{kirkpatrick2017overcoming}, a regularization method that selectively updates less critical weights to preserve prior task knowledge; (3) \textbf{ER}~\cite{chaudhry2019tiny}, a rehearsal-based approach with a 1000-trajectory memory buffer to retain prior task samples; (4) \textbf{BUDS}~\cite{zhu2022bottom}, a hierarchical policy method using multitask skill discovery; (5) \textbf{LOTUS}~\cite{wan2024lotus}, a hierarchical imitation learning framework with open-vocabulary vision and experience replay for unsupervised skill discovery; and (6) \textbf{M2Distill}~\cite{roy2024m2distill}, a multi-modal distillation method maintaining consistent latent spaces across vision, language, and action modalities. Baseline results are sourced from M2Distill~\cite{roy2024m2distill}.

\subsection{Datasets}
\label{sec:datasets}

For our evaluations, we utilize the LIBERO benchmark~\cite{liu2024libero}, designed for lifelong imitation learning in robotic manipulation. LIBERO encompasses diverse, language-conditioned tasks with varied objects, sparse rewards, and long-horizon objectives, making it ideal for assessing continual learning. We focus on three task suites, each comprising 10 sequential tasks: LIBERO-OBJECT, LIBERO-GOAL, and LIBERO-SPATIAL. These suites evaluate knowledge transfer in distinct dimensions: object-specific declarative knowledge, procedural goal-oriented behaviors, and spatial relational understanding, respectively. LIBERO-OBJECT tasks require continual learning to manipulate distinct objects. LIBERO-GOAL tasks involve identical objects with different spatial arrangements and task objectives, necessitating adaptation of motions and behaviors. LIBERO-SPATIAL tasks challenge the robot to distinguish identical objects based solely on spatial context, demanding persistent learning of spatial relationships. This setup rigorously tests our subspace representation distillation framework's ability to preserve geometric structures and action distributions across tasks.

\smallskip
\subsection{Evaluation Metrics}

To assess policy performance in lifelong imitation learning for robotic manipulation tasks, we adopt three key metrics: Forward Transfer (FWT), Negative Backward Transfer (NBT), and Area Under the Success Rate Curve (AUC), following established protocols~\cite{liu2024libero, wan2024lotus}. These metrics, based on task success rates, offer a robust evaluation compared to training loss, which is less reliable for manipulation tasks. FWT, defined as $\text{FWT} = \frac{1}{K} \sum_{k=1}^K \text{FWT}_k$ where $\text{FWT}_k = \frac{1}{51} \sum_{e=0}^{50} c_{k,k,e}$, measures adaptation to new tasks using prior knowledge, with higher values indicating better transfer. NBT, $\text{NBT} = \frac{1}{K} \sum_{k=1}^K \frac{1}{K - k} \sum_{\tau=k+1}^K (c_{k,k} - c_{\tau,k})$, quantifies forgetting of prior tasks, with lower values reflecting stronger retention. AUC, $\text{AUC} = \frac{1}{K} \sum_{k=1}^K \frac{1}{K - k + 1} \left( \text{FWT}_k + \sum_{\tau=k+1}^K c_{\tau,k} \right)$, assesses overall task performance, with higher scores denoting superior success. Here, $K$ is the number of tasks, $c_{k,k,e}$ is the success rate of task $k$ at epoch $e$, and $c_{\tau,k}$ is the success rate of task $k$ after training on task $\tau$~\cite{liu2024libero}.

\begin{figure*}[t!]
  \centering
  \begin{subfigure}[b]{0.47\textwidth}
    \centering
    \includegraphics[width=\textwidth, height=5.5cm]{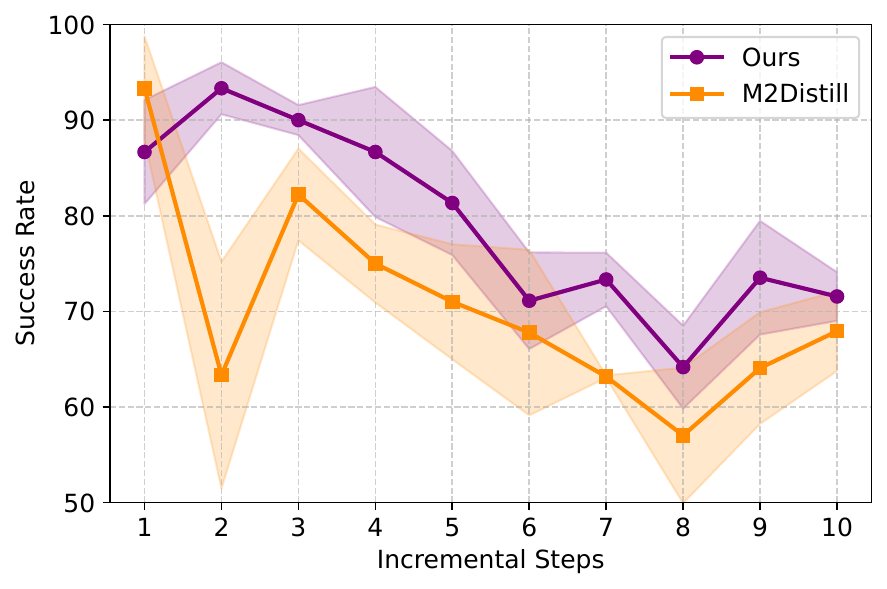}
    \caption{LIBERO-OBJECT}
    \label{fig:image1}
  \end{subfigure}
  \hfill
  \begin{subfigure}[b]{0.47\textwidth}
    \centering
    \includegraphics[width=\textwidth, height=5.5cm]{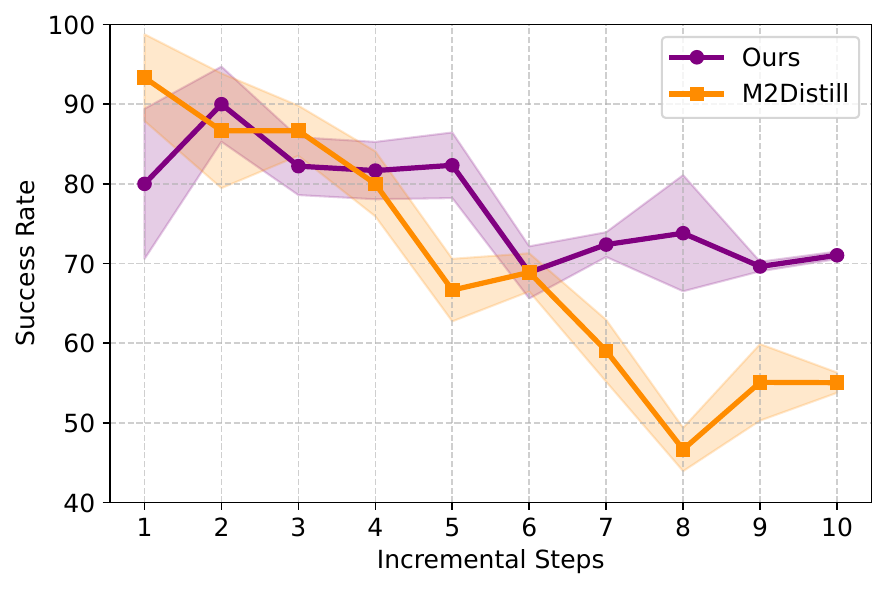}
    \caption{LIBERO-GOAL}
    \label{fig:image2}
  \end{subfigure}
  \caption{Average success rate across incremental tasks on (a) LIBERO-OBJECT and (b) LIBERO-GOAL (Higher numbers are better).}
  \label{fig:inc_success_rate}
  \vspace{-1em}
\end{figure*}

\section{Experimental evaluation}

\noindent
\textbf{Comparison to SOTA.}
We evaluate our Subspace Representation Distillation framework, which aligns low-rank feature projections and employs confidence-guided policy distillation with top-\( K \) samples, by comparing it against state-of-the-art (SOTA) methods on the LIBERO benchmark~\cite{liu2024libero}, comprising three task suites (LIBERO-OBJECT, LIBERO-GOAL, LIBERO-SPATIAL), each with 10 sequential robotic manipulation tasks. Following prior work, we report FWT, NBT, and AUC, with results averaged over three random seeds (see Table~\ref{tab:benchmark}). Across all benchmarks, \coolname{} achieves the best or competitive performance compared to strong baselines such as LOTUS and M2Distill. On \textbf{LIBERO-OBJECT}, \coolname{} obtains the highest FWT (81.0\% \(\pm\),  2.0), AUC (73.0\% \(\pm\) 3.0), while also maintaining the lowest NBT (8.0\% \(\pm\) 1.0), outperforming M2Distill by +6\% in FWT and +4\% in AUC. On \textbf{LIBERO-GOAL}, which presents more challenging goal-conditioned tasks, \coolname{} surpasses prior methods by a large margin, achieving an FWT of 78.0\% \(\pm\) 1.0 and an NBT of only 9.0\% \(\pm\) 5.0; in contrast, LOTUS and M2Distill suffer significantly higher forgetting (30.0\% \(\pm\) 1.0 and 20.0\% \(\pm\) 3.0, respectively). Finally, on \textbf{LIBERO-SPATIAL}, \coolname{} reaches the best AUC (66.0\% \(\pm\) 2.0) while keeping NBT low (8.0\% \(\pm\) 5.0), outperforming M2Distill by +5\% in AUC with comparable FWT. These results demonstrate \coolname{}’s superior adaptation (FWT) and retention (NBT), driven by its low-rank subspace alignment and confidence-guided policy distillation, which preserve intrinsic task manifolds and robust action distributions across diverse LIL scenarios.

\noindent
\textbf{Success Rate Analysis.} We evaluate our subspace representation distillation and M2Distill baseline across ten incremental steps on the LIBERO-OBJECT and LIBERO-GOAL datasets to assess robustness in continual learning and present the result in \figref{fig:inc_success_rate}. On LIBERO-OBJECT, our method achieved success rates from $64.0\% \pm 4.0\%$ to a peak of $93.0\% \pm 3.0\%$ at step 2, staying above $70.0\%$, while M2Distill started at $93.0\% \pm 5.0\%$ but dropped to $68.0\% \pm 4.0\%$ by step 10, with variability like $63.0\% \pm 12.0\%$ at step 2. On LIBERO-GOAL, our approach maintained rates from $69.0\% \pm 3.0\%$ to $90.0\% \pm 5.0\%$, peaking at step 2 from an $80.0\% \pm 9.0\%$ start, whereas M2Distill fell from $93.0\% \pm 5.0\%$ to $55.0\% \pm 1.0\%$ by step 10, hitting $47.0\% \pm 3.0\%$ at step 8. These results highlight our method's superior consistency and knowledge retention over M2Distill in continual learning.

\begin{figure*}[t!]
  \centering
  \begin{subfigure}[b]{0.47\textwidth}
    \centering
    \includegraphics[width=\textwidth, height=5.5cm]{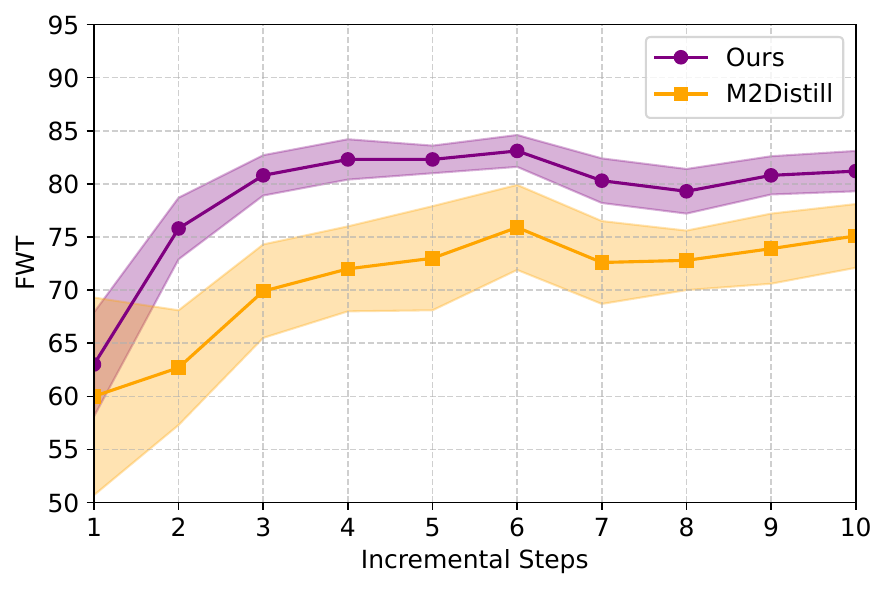}
    \caption{LIBERO-OBJECT}
    \label{fig:image1}
  \end{subfigure}
  \hfill
  \begin{subfigure}[b]{0.47\textwidth}
    \centering
    \includegraphics[width=\textwidth, height=5.5cm]{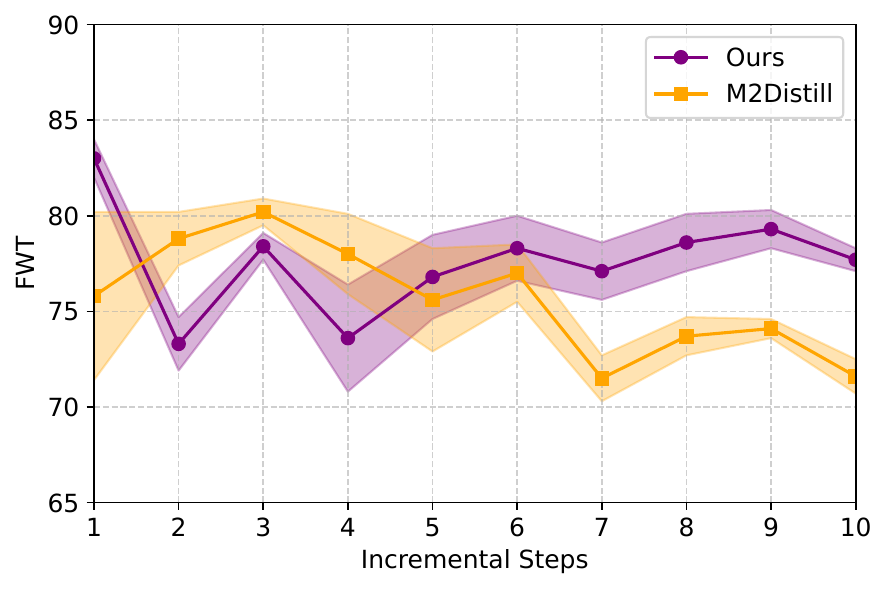}
    \caption{LIBERO-GOAL}
    \label{fig:image2}
  \end{subfigure}
  \caption{Forward Transfer (FWT) across incremental tasks on (a) LIBERO-OBJECT and (b) LIBERO-GOAL (Higher numbers are better).}
  \label{fig:inc_fwt}
  \vspace{-1.5em}
\end{figure*}

\noindent
\textbf{FWT Analysis.} We assess the Forward Transfer (FWT) of our subspace representation distillation strategy and the M2Distill baseline across 10 incremental steps on the LIBERO-OBJECT and LIBERO-GOAL datasets to evaluate their ability to utilize prior knowledge for new tasks in continual learning. Results are illustrated in \figref{fig:inc_fwt}. FWT quantifies performance improvement on new tasks from prior learning, with our method achieving robust FWT values of $63.0\% \pm 5.0\%$ to $83.0\% \pm 2.0\%$ on LIBERO-OBJECT, peaking at step 6 with standard errors of $1.0\%$ to $5.0\%$, compared to M2Distill's $60.0\% \pm 9.0\%$ to $75.0\% \pm 3.0\%$, peaking at step 10 with higher variability up to $9.0\%$. On LIBERO-GOAL, our approach maintained stable FWT values of $73.0\% \pm 1.0\%$ to $83.0\% \pm 1.0\%$, peaking at step 1 with errors of $1.0\%$ to $3.0\%$, while M2Distill ranged from $72.0\% \pm 1.0\%$ to $80.0\% \pm 1.0\%$, peaking at step 3 with errors up to $4.0\%$. These findings highlight our strategy's consistent superiority over M2Distill in knowledge transfer, enhancing continual learning performance in LIL.

\begin{figure}[t!]
\hspace{-1em}
\centering
\includegraphics[width=.47\textwidth, height=5.5cm, scale=1]{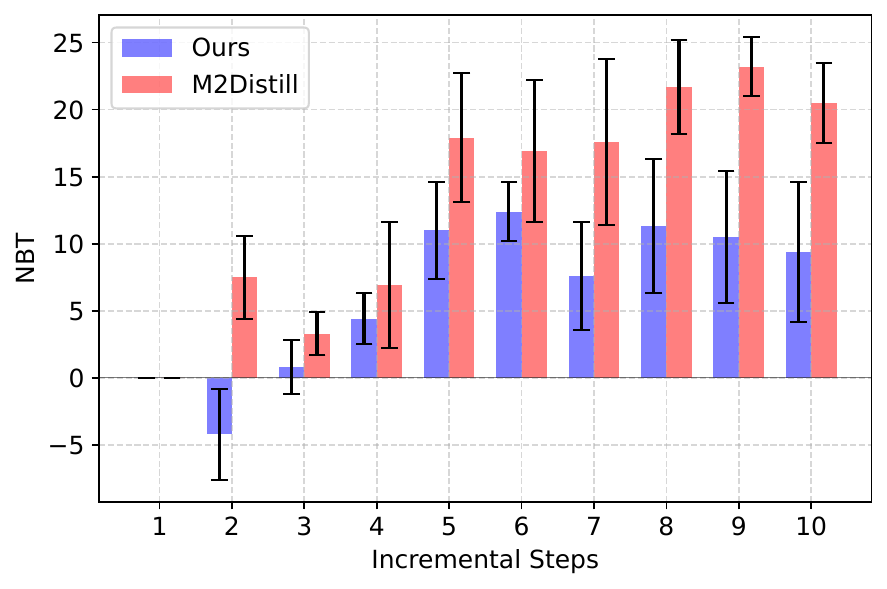}
\caption{Negative Backward Transfer (NBT) across incremental tasks on LIBERO-GOAL (lower is better). } 
\label{fig:inc_nbt} 
\vspace{-2.0em}
\end{figure}

\noindent
\textbf{NBT Analysis.} We assess the Negative Backward Transfer (NBT) of our subspace representation distillation strategy and the M2Distill baseline across ten incremental steps on the LIBERO-GOAL dataset to evaluate forgetting in continual learning. Results are depicted in \figref{fig:inc_nbt}. NBT measures performance decline on prior tasks with new task introduction. Our method shows NBT values from $-4.0 \pm 3.0$ at step 2 to $12.0 \pm 2.0$ at step 6, featuring modest positive values (1.0 to 12.0) and errors of 2.0\% to 5.0\%, indicating minimal forgetting and a gain at step 2. Conversely, M2Distill displayed higher NBT values, ranging from $3.0 \pm 2.0$ at step 3 to $23.0 \pm 2.0$ at step 9, with errors up to 6.0\%, reflecting greater forgetting and a peak at step 9. These findings highlight our strategy's superior stability in reducing forgetting compared to M2Distill on the LIBERO-GOAL dataset.

\begin{figure}[t!]
\hspace{-1em}
\centering
\includegraphics[width=.49\textwidth, scale=1]{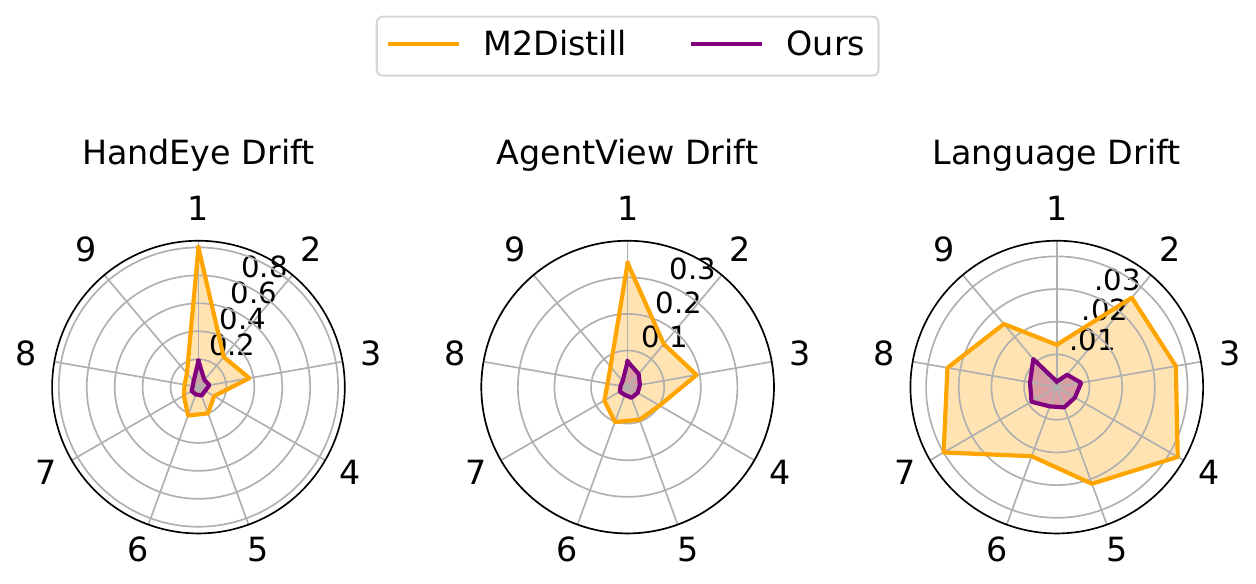}
\caption{Drift analysis across incremental tasks for different modalities on LIBERO-OBJECT. The radial axes (1-9) indicate the incremental steps in LIL. Lower representation drift indicates better preservation of previously learned knowledge across tasks in lifelong imitation learning.
} 
\label{fig:inc_drift} 
\vspace{-1.5em}
\end{figure}

\smallskip
\noindent
\textbf{Drift Analysis.} We evaluate representation drift, defined as the change in learned feature embeddings between two consecutive tasks, measured as the mean distance between their representations. The evaluation is performed across language, wrist-mounted camera (HandEye), and overhead camera (AgentView) embedding spaces over successive incremental learning steps on the LIBERO-OBJECT task suite with a fixed random seed of 100. Overall, we observe that language embeddings remain the most stable across incremental steps, while hand-eye and agent-view embeddings exhibit higher drift. As shown in \figref{fig:inc_drift}, our proposed subspace representation distillation method significantly outperforms M2Distill in preserving feature stability across all modalities by maintaining similar subspaces. In the language embedding space, \coolname{} reduces drift by more than 75\%, demonstrating strong preservation of semantic representations. In the visual modalities, it effectively suppresses the large drift spikes observed in M2Distill, keeping drift in the HandEye embedding space below 0.5 compared to the baseline peak of over 2.7. For AgentView, \coolname{} again outperforms with a mean drift of 0.095 against 0.356 for M2Distill. Overall, these results demonstrate that subspace representation distillation effectively reduces drift across modalities, mitigating catastrophic forgetting in lifelong imitation learning.

\section{Ablation Study} 

\noindent
\textbf{Influence of each loss component.}
We further investigate the contribution of individual loss components through an ablation study on the LIBERO-GOAL task suite, with results summarized in Table~\ref{tab:ablation_study}. The removal of the text loss $\mathcal{L}_{\text{text}}$ yields a slight reduction in AUC (0.72 $\rightarrow$ 0.69), indicating that language alignment provides auxiliary benefits but is not the dominant factor. In contrast, excluding the image loss $\mathcal{L}_{\text{image}}$ leads to a substantial degradation in performance, with AUC dropping from 0.72 to 0.57 and NBT increasing from 0.09 to 0.20, thereby underscoring the critical role of visual representation preservation in mitigating forgetting. Eliminating the action loss $\mathcal{L}_{\text{policy}}$ results in decreased forward transfer (0.78 $\rightarrow$ 0.70), despite a marginal improvement in NBT, suggesting that action alignment is essential for retaining transferable policies across tasks. Finally, the exclusion of the extra-modality loss $\mathcal{L}_{\text{extra}}$ produces a minor decline in both FWT and AUC, confirming its complementary role in enhancing representational consistency. Collectively, these findings demonstrate that while all components contribute to the robustness of the proposed framework, image feature alignment emerges as the most influential factor for effective lifelong imitation learning.

\begin{table}[b!]
\vspace{-1em}
\centering
\caption{\label{tab:ablation_study} Ablation studies on the contribution of each component in our method. Experiments were performed on the LIBERO-GOAL task suite. Reported results are the mean across three different seed values.
}
\vspace{-.5em}
\resizebox{.49\textwidth}{!}{%
\begin{tabular}{cccc|ccc}
\multicolumn{4}{c}{}                                                               & \multicolumn{3}{c}{}                                               \\ \hline
\multirow{2}{*}{$\mathcal{L}_{\text{text}}$} & \multirow{2}{*}{$\mathcal{L}_{\text{image}}$} & \multirow{2}{*}{$\mathcal{L}_{\text{action}}$} & \multirow{2}{*}{$\mathcal{L}_{\text{extra}}$} & \multicolumn{3}{c}{LIBERO-GOAL} \\ 
&            &            &         & FWT $\uparrow$ & NBT $\downarrow$ & AUC $\uparrow$  \\ \hline
\cmark   &  \cmark   &   \cmark  &   \cmark  &   0.78    &   0.09    &   0.72    \\
\xmark   &  \cmark   &   \cmark  &   \cmark  &   0.78    &   0.11    &   0.69    \\
\cmark   &  \xmark   &   \cmark  &   \cmark  &   0.67    &   0.20    &   0.57    \\
\cmark   &  \cmark   &   \xmark  &   \cmark  &   0.70    &   0.02    &    0.70   \\
\cmark   &  \cmark   &   \cmark  &   \xmark  &   0.74    &   0.03    &   0.71    \\ 
\hline

\end{tabular} }
\end{table}

\noindent
\textbf{Dimensionality of Subspace Representation Distillation.}
We investigate the impact of subspace dimensionality in \coolname{} by varying the projection rank $r$ on the LIBERO-GOAL task suite (Table~\ref{tab:ablation_subspace}). The results demonstrate that the choice of rank has a significant effect on both knowledge transfer and forgetting. Although the full-rank ($r=64$) baseline attains competitive FWT, it suffers from higher forgetting and a lower AUC. In contrast, the \(75\%\)-rank configuration (\(r=48\)) achieves the best overall performance, with the highest AUC (71.0\% \(\pm\) 1.0) and FWT (78.0\% \(\pm\) 1.0), outperforming the full-rank \coolname{} by 5\% in AUC and 7\% in NBT. This suggests that a moderately compressed subspace effectively retains transferable information while mitigating redundancy. However, further reducing the rank (\(r=32, 16\)) causes noticeable information loss, leading to degraded FWT and AUC. Following these results, we set the \(75\%\)-rank as the default in all subsequent experiments.

\begin{figure}[t!]
\hspace{-1em}
\centering
\includegraphics[width=.47\textwidth, height=5.25cm, scale=1]{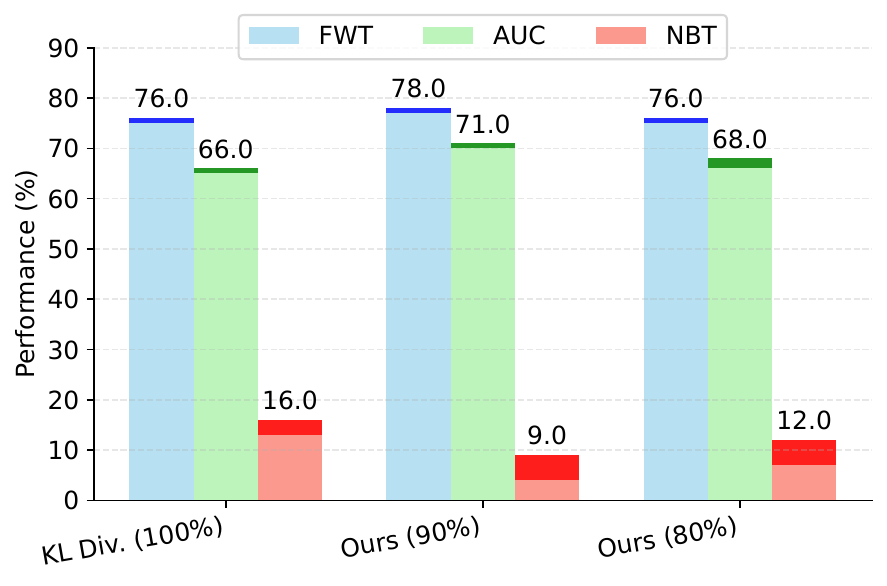}
\caption{Ablation study on policy distillation strategies using LIBERO-GOAL task suite. Bar represents mean across three random seeds while standard error is represented with darker color. Our top-\( M \) confidence-weighted approach outperforms baselines in FWT, NBT and AUC.
} 
\label{fig:ablation_policy} 
\end{figure}

\begin{table}[t]
\centering
\caption{Ablation study on the dimensionality/rank of subspace representation distillation strategies using LIBERO-GOAL task suite. Our proposed \coolname{} loss with \(75\%\)-rank performs best in FWT, NBT and AUC.}
\label{tab:ablation_subspace}
\resizebox{.49\textwidth}{!}{%

\begin{tabular}{lccc}
\toprule
\textbf{Rank of Subspace} & \textbf{FWT ($\uparrow$)} & \textbf{BWT ($\downarrow$)} & \textbf{AUC ($\uparrow$)} \\
\midrule
\coolname{} (Full-Rank) & 76.0 ($\pm$ 1.0) & 16.0 ($\pm$ 8.0) & 66.0 ($\pm$ 5.0)\\
\coolname{} (Rank \(=48\)) & \textbf{78.0 ($\pm$ 1.0)} & \textbf{9.0 ($\pm$ 5.0)} & \textbf{72.0 ($\pm$ 1.0)} \\
\coolname{} (Rank \(=32\)) & 74.0 ($\pm$ 1.0) & 6.0 ($\pm$ 4.0) & 71.0 ($\pm$ 2.0) \\
\coolname{} (Rank \(=16\)) & 76.0 ($\pm$ 3.0) & 9.0 ($\pm$ 4.0) & 69.0 ($\pm$ 2.0) \\
\bottomrule
\end{tabular}
}
\vspace{-1.5em}
\end{table}

\noindent
\textbf{Policy Distillation with Top-\( M \) Confident Samples.}
We evaluate the effectiveness of our confidence-based policy distillation strategy, which selects the top \( M \) samples with the highest log-probability scores under the previous policy \(\pi^{k-1}\). Experiments are conducted on the LIBERO-GOAL suite, comprising 10 consecutive robotic manipulation tasks. The policy distillation loss, \(\mathcal{L}_{\text{policy}}\), complements subspace representation distillation by enforcing alignment of action distributions. As shown in \figref{fig:ablation_policy}, our method—combining low-rank subspace projection with top-\( M = \lfloor 0.9B \rfloor \) confident samples—achieves the best results, with an AUC of 71.0\% \(\pm\) 1.0, FWT of 78.0\% \(\pm\) 1.0, and NBT of 9.0\% \(\pm\) 5.0. This outperforms standard KL divergence using all samples by 5\% in AUC and 7\% in NBT. In contrast, a variant with \( M = \lfloor 0.8B \rfloor \) samples shows reduced performance (AUC: 68.0\% \(\pm\) 2.0, NBT: 12.0\% \(\pm\) 5.0), suggesting that \( M = \lfloor 0.9B \rfloor \) provides a good balance between sample diversity and confidence. Overall, this confidence-based top-\( M \) selection enhances robustness by prioritizing reliable action distributions, complementing geometric preservation in feature distillation, and improving both stability and generalization across sequential tasks. %

\section{Conclusion}
\label{sec:conclusion}

This work presented \textit{\coolname{}}, a modality-aware framework for lifelong imitation learning that explicitly leverages subspace geometry to mitigate catastrophic forgetting. By aligning features of two consecutive policies through low-rank projections derived via singular value decomposition, \textit{\coolname{}} preserves the intrinsic low-dimensional structure of multimodal task embeddings more effectively than conventional feature-space distillation. Complementary to this, the proposed confidence-guided policy distillation selectively emphasizes high-confidence action samples, yielding a more stable and reliable transfer of behavioral priors. On the LIBERO benchmark, \textit{\coolname{}} achieves AUC scores of 73.0\%, 72.0\%, and 66.0\% on LIBERO-OBJECT, LIBERO-GOAL, and LIBERO-SPATIAL, outperforming state-of-the-art M2Distill by 4--15\% in AUC and reducing negative backward transfer by 12\%. These findings highlight both the empirical benefits and the theoretical advantages of subspace-level alignment for continual representation learning. Future research will explore \textit{\coolname{}}’s application to long-horizon incremental task learning and addressing extended task sequences for real-world adaptability.

 \balance{}
\bibliographystyle{IEEEtran}
\bibliography{ref}

@IEEEtranBSTCTL{IEEEexample:BSTcontrol,
CTLuse_forced_etal       = "yes",
CTLmax_names_forced_etal = "3",
CTLnames_show_etal       = "2" }

@article{liu2024libero,
  title={Libero: Benchmarking knowledge transfer for lifelong robot learning},
  author={Liu, Bo and Zhu, Yifeng and Gao, Chongkai and Feng, Yihao and Liu, Qiang and Zhu, Yuke and Stone, Peter},
  journal={Advances in Neural Information Processing Systems},
  volume={36},
  year={2024}
}

@inproceedings{wan2024lotus,
  title={Lotus: Continual imitation learning for robot manipulation through unsupervised skill discovery},
  author={Wan, Weikang and Zhu, Yifeng and Shah, Rutav and Zhu, Yuke},
  booktitle={2024 IEEE International Conference on Robotics and Automation (ICRA)},
  pages={537--544},
  year={2024},
  organization={IEEE}
}

@inproceedings{roy2024m2distill,
  title={{M2Distill}: Multi-modal distillation for lifelong imitation learning},
  author={Roy, Kaushik and Dissanayake, Akila and Tidd, Brendan and Moghadam, Peyman},
  booktitle={2025 IEEE International Conference on Robotics and Automation (ICRA)},
  pages={1429--1435},
  year={2025},
}

@article{MAHMOODI2025,
title = {Flashbacks to harmonize stability and plasticity in continual learning},
journal = {Neural Networks},
volume = {190},
pages = {107616},
year = {2025},
author = {Leila Mahmoodi and Peyman Moghadam and Munawar Hayat and Christian Simon and Mehrtash Harandi}
}

@article{haldar2023polytask,
  title={Polytask: Learning unified policies through behavior distillation},
  author={Haldar, Siddhant and Pinto, Lerrel},
  journal={arXiv preprint arXiv:2310.08573},
  year={2023}
}

@inproceedings{hershey2007approximating,
  title={Approximating the {Kullback} {Leibler} divergence between {Gaussian} mixture models},
  author={Hershey, John R and Olsen, Peder A},
  booktitle={2007 IEEE International Conference on Acoustics, Speech and Signal Processing-ICASSP'07},
  volume={4},
  pages={IV--317},
  year={2007},
  organization={IEEE}
}

@article{zhu2022bottom,
  title={Bottom-up skill discovery from unsegmented demonstrations for long-horizon robot manipulation},
  author={Zhu, Yifeng and Stone, Peter and Zhu, Yuke},
  journal={IEEE Robotics and Automation Letters},
  volume={7},
  number={2},
  pages={4126--4133},
  year={2022},
  publisher={IEEE}
}

@article{chaudhry2019tiny,
  title={On tiny episodic memories in continual learning},
  author={Chaudhry, Arslan and Rohrbach, Marcus and Elhoseiny, Mohamed and Ajanthan, Thalaiyasingam and Dokania, Puneet K and Torr, Philip HS and Ranzato, Marc'Aurelio},
  journal={arXiv preprint arXiv:1902.10486},
  year={2019}
}

@inproceedings{bain1995framework,
  title={A Framework for Behavioural Cloning.},
  author={Bain, Michael and Sammut, Claude},
  booktitle={Machine Intelligence 15},
  pages={103--129},
  year={1995}
}

@inproceedings{he2016deep,
  title={Deep residual learning for image recognition},
  author={He, Kaiming and Zhang, Xiangyu and Ren, Shaoqing and Sun, Jian},
  booktitle={Proceedings of the IEEE conference on computer vision and pattern recognition},
  pages={770--778},
  year={2016}
}

@article{kirkpatrick2017overcoming,
  title={Overcoming catastrophic forgetting in neural networks},
  author={Kirkpatrick, James and Pascanu, Razvan and Rabinowitz, Neil and Veness, Joel and Desjardins, Guillaume and Rusu, Andrei A and Milan, Kieran and Quan, John and Ramalho, Tiago and Grabska-Barwinska, Agnieszka and others},
  journal={Proceedings of the National Academy of Sciences},
  volume={114},
  number={13},
  pages={3521--3526},
  year={2017}
}

@inproceedings{gao2021cril,
  title={{CRIL}: Continual robot imitation learning via generative and prediction model},
  author={Gao, Chongkai and Gao, Haichuan and Guo, Shangqi and Zhang, Tianren and Chen, Feng},
  booktitle={2021 IEEE/RSJ International Conference on Intelligent Robots and Systems (IROS)},
  pages={6747--5754},
  year={2021}
}

@article{stepputtis2020language,
  title={Language-conditioned imitation learning for robot manipulation tasks},
  author={Stepputtis, Simon and Campbell, Joseph and Phielipp, Mariano and Lee, Stefan and Baral, Chitta and Ben Amor, Heni},
  journal={Advances in Neural Information Processing Systems},
  volume={33},
  pages={13139--13150},
  year={2020}
}

@inproceedings{xie2024decomposing,
  title={Decomposing the generalization gap in imitation learning for visual robotic manipulation},
  author={Xie, Annie and Lee, Lisa and Xiao, Ted and Finn, Chelsea},
  booktitle={2024 IEEE International Conference on Robotics and Automation (ICRA)},
  pages={3153--3160},
  year={2024},
  organization={IEEE}
}

@inproceedings{jang2022bc,
  title={Bc-z: Zero-shot task generalization with robotic imitation learning},
  author={Jang, Eric and Irpan, Alex and Khansari, Mohi and Kappler, Daniel and Ebert, Frederik and Lynch, Corey and Levine, Sergey and Finn, Chelsea},
  booktitle={Conference on Robot Learning},
  pages={991--1002},
  year={2022},
  organization={PMLR}
}

@inproceedings{roy2023l3dmc,
  title={{L3DMC}: Lifelong Learning using Distillation via Mixed-Curvature Space},
  author={Roy, Kaushik and Moghadam, Peyman and Harandi, Mehrtash},
  booktitle={International Conference on Medical Image Computing and Computer-Assisted Intervention},
  pages={123--133},
  year={2023}
}

@article{roy2023subspace,
  title={Subspace Distillation for Continual Learning},
  author={Roy, Kaushik and Simon, Christian and Moghadam, Peyman and Harandi, Mehrtash},
  journal={Neural Networks},
  volume={167},
  pages={65--79},
  year={2023},
  publisher={Elsevier}
}

@article{hussein2017imitation,
  title={Imitation learning: A survey of learning methods},
  author={Hussein, Ahmed and Gaber, Mohamed Medhat and Elyan, Eyad and Jayne, Chrisina},
  journal={ACM Computing Surveys (CSUR)},
  volume={50},
  number={2},
  pages={1--35},
  year={2017},
  publisher={ACM New York, NY, USA}
}

@article{billard2016learning,
  title={Learning from humans},
  author={Billard, Aude G and Calinon, Sylvain and Dillmann, R{\"u}diger},
  journal={Springer handbook of robotics},
  pages={1995--2014},
  year={2016},
  publisher={Springer}
}

@article{schaal1996learning,
  title={Learning from demonstration},
  author={Schaal, Stefan},
  journal={Advances in neural information processing systems},
  volume={9},
  year={1996}
}

@article{french1999catastrophic,
  title={Catastrophic forgetting in connectionist networks},
  author={French, Robert M},
  journal={Trends in cognitive sciences},
  volume={3},
  number={4},
  pages={128--135},
  year={1999}
}

@inproceedings{shmelkov2017incremental,
  title={Incremental learning of object detectors without catastrophic forgetting},
  author={Shmelkov, Konstantin and Schmid, Cordelia and Alahari, Karteek},
  booktitle={Proceedings of the IEEE international conference on computer vision},
  pages={3400--3409},
  year={2017}
}

@inproceedings{rebuffi2017icarl,
  title={{icarl}: Incremental classifier and representation learning},
  author={Rebuffi, Sylvestre-Alvise and Kolesnikov, Alexander and Sperl, Georg and Lampert, Christoph H},
  booktitle={Proceedings of the IEEE conference on Computer Vision and Pattern Recognition},
  pages={2001--2010},
  year={2017}
}

@article{ahn2019uncertainty,
  title={Uncertainty-based continual learning with adaptive regularization},
  author={Ahn, Hongjoon and Cha, Sungmin and Lee, Donggyu and Moon, Taesup},
  journal={Advances in neural information processing systems},
  volume={32},
  year={2019}
}

@inproceedings{douillard2022dytox,
  title={Dytox: Transformers for continual learning with dynamic token expansion},
  author={Douillard, Arthur and Ram{\'e}, Alexandre and Couairon, Guillaume and Cord, Matthieu},
  booktitle={Proceedings of the IEEE/CVF Conference on Computer Vision and Pattern Recognition},
  pages={9285--9295},
  year={2022}
}

@article{rolnick2019experience,
  title={Experience replay for continual learning},
  author={Rolnick, David and Ahuja, Arun and Schwarz, Jonathan and Lillicrap, Timothy and Wayne, Gregory},
  journal={Advances in neural information processing systems},
  volume={32},
  year={2019}
}

@inproceedings{yoon2017lifelong,
  title     = {Lifelong Learning with Dynamically Expandable Networks},
  author    = {Yoon, Jaehong and Yang, Eunho and Lee, Jeongtae and Hwang, Sung Ju},
  booktitle = {International Conference on Learning Representations (ICLR)},
  year      = {2018},
  eprint    = {1708.01547},
  archivePrefix = {arXiv},
  primaryClass = {cs.LG},
}

@article{shin2017continual,
  title={Continual learning with deep generative replay},
  author={Shin, Hanul and Lee, Jung Kwon and Kim, Jaehong and Kim, Jiwon},
  journal={Advances in neural information processing systems},
  volume={30},
  year={2017}
}

@inproceedings{huang2023knowledge,
  title={Knowledge transfer in incremental learning for multilingual neural machine translation},
  author={Huang, Kaiyu and Li, Peng and Ma, Jin and Yao, Ting and Liu, Yang},
  booktitle={Proceedings of the 61st Annual Meeting of the Association for Computational Linguistics)},
  pages={15286--15304},
  year={2023}
}

@article{zhang2025t2s,
  title={{T2S}: Tokenized Skill Scaling for Lifelong Imitation Learning},
  author={Zhang, Hongquan and Gong, Jingyu and Zhang, Zhizhong and Tan, Xin and Qu, Yanyun and Xie, Yuan},
  journal={arXiv preprint arXiv:2508.01167},
  year={2025}
}

@article{xu2025speci,
  title={{SPECI}: Skill Prompts based Hierarchical Continual Imitation Learning for Robot Manipulation},
  author={Xu, Jingkai and Nie, Xiangli},
  journal={arXiv preprint arXiv:2504.15561},
  year={2025}
}

@article{hinton2015distilling,
  title={Distilling the knowledge in a neural network},
  author={Hinton, Geoffrey and Vinyals, Oriol and Dean, Jeff and others},
  journal={arXiv preprint arXiv:1503.02531},
  volume={2},
  number={7},
  year={2015}
}

@inproceedings{cheng2021nbnet,
  title={Nbnet: Noise basis learning for image denoising with subspace projection},
  author={Cheng, Shen and Wang, Yuzhi and Huang, Haibin and Liu, Donghao and Fan, Haoqiang and Liu, Shuaicheng},
  booktitle={Proceedings of the IEEE/CVF Conference on Computer Vision and Pattern Recognition},
  pages={4896--4906},
  year={2021}
}

@article{wall2003singular,
  title={Singular value decomposition and principal component analysis},
  author={Wall, Michael E and Rechtsteiner, Andreas and Rocha, Luis M},
  journal={A practical approach to microarray data analysis},
  pages={91--109},
  year={2003},
  publisher={Springer}
}

@article{eckart1936approximation,
  title={The approximation of one matrix by another of lower rank},
  author={Eckart, Carl and Young, Gale},
  journal={Psychometrika},
  volume={1},
  number={3},
  pages={211--218},
  year={1936},
  publisher={Springer}
}

@inproceedings{radford2021learning,
  title={Learning transferable visual models from natural language supervision},
  author={Radford, Alec and Kim, Jong Wook and Hallacy, Chris and Ramesh, Aditya and Goh, Gabriel and Agarwal, Sandhini and Sastry, Girish and Askell, Amanda and Mishkin, Pamela and Clark, Jack and others},
  booktitle={International conference on machine learning},
  pages={8748--8763},
  year={2021},
  organization={PmLR}
}

@article{kullback1951information,
  title={On information and sufficiency},
  author={Kullback, Solomon and Leibler, Richard A},
  journal={The annals of mathematical statistics},
  volume={22},
  number={1},
  pages={79--86},
  year={1951},
  publisher={JSTOR}
}

\end{document}